%% file: main.tex
\documentclass[letterpaper]{article}
\usepackage{aaai2027}
\usepackage[hyphens]{url}
\usepackage{graphicx}
\usepackage{natbib}
\usepackage{caption}
\usepackage{algorithm}
\usepackage{algorithmic}
\usepackage{amsmath}
\usepackage{amssymb}
\usepackage{booktabs}
\usepackage{tabularx}
\usepackage{multirow}
\usepackage{colortbl}
\usepackage{xspace}

\newcommand{\method}{\textbf{\texttt{DynaKRAG}}\xspace}
\newcommand{\best}[1]{\textbf{#1}}
\newcommand{\second}[1]{\underline{#1}}
\definecolor{ourblue}{RGB}{235,235,235}
\makeatletter
\newcommand{\fullrowshade}[1]{%
  \smash{\rlap{%
    \color{#1}%
    \rule[-\dp\@arstrutbox]%
      {\textwidth}%
      {\dimexpr\ht\@arstrutbox+\dp\@arstrutbox\relax}%
  }}%
}
\makeatother

\title{\method: A Unified Framework for Learnable Evidence Control in Multi-Hop Retrieval-Augmented Generation}
\author{
Chenyu Zhou\textsuperscript{\rm 1},
Yaqi Wu\textsuperscript{\rm 1}\equalcontrib,
Xiaolei Guo\textsuperscript{\rm 1}\equalcontrib,
Jiaqi Huang\textsuperscript{\rm 1},
Xianfa Zhang\textsuperscript{\rm 2},
Junxu Zhang\textsuperscript{\rm 2},
Zhuo Yu\textsuperscript{\rm 2},
Zhubo Shi\textsuperscript{\rm 3},
Jianghao Lin\textsuperscript{\rm 1}\corresponding,
Dongdong Ge\textsuperscript{\rm 1}
}

\affiliations{
\textsuperscript{\rm 1}Shanghai Jiao Tong University\\
\textsuperscript{\rm 2}Shanghai Aircraft Manufacturing Co., Ltd.\\
\textsuperscript{\rm 3}Tongji University\\
\{chenyuzhou, linjianghao, ddge\}@sjtu.edu.cn
}

\begin{document}

\maketitle

\input{sections/00_abstract}

\input{sections/01_introduction}
\input{sections/02_related_work}
\input{sections/03_method}
\input{sections/04_experiments}
\input{sections/07_conclusion}

\bibliography{references}

\end{document}

%% file: sections/00_abstract.tex
\begin{abstract}
Multi-hop retrieval-augmented generation (RAG) acquires evidence sequentially, with each document contributing supporting facts, bridge entities, query refinements, or sufficient evidence for answering. Evidence acquisition can involve iterative retrieval, query reformulation, evidence assessment, and sufficiency checking. We introduce \method, a unified evidence-action framework that learns a shared state-conditioned policy for coordinating these operations. At each step, a deterministic validity layer constructs the executable action set, a learned continuation gate selects between answer generation and further evidence acquisition, and a learned advantage scorer ranks feasible evidence operations by their predicted gain relative to immediate answer generation. The selected operation updates the shared state and may enable additional operations. Across HotpotQA, 2Wiki, and MuSiQue with Qwen2.5-7B, GPT-4o-mini, and Llama-3.1-8B, \method ranks first among the compared methods in both EM and F1 for all nine dataset--backbone pairs. Relative to matched-backbone baseline method, \method improves F1 in every pair while achieving total-token efficiency gains of 10.1--34.3\% and retrieval-call efficiency gains of 15.1--43.4\%, establishing Pareto dominance under these measures. With Qwen2.5-7B, terminal evidence compression further improves answer quality across all three datasets while reducing the context passed to final answer generation by 54.4\%--71.5\%. These results demonstrate that unified, state-conditioned evidence control supports strong answer quality, efficient retrieval, and compact answer-generation contexts.
\end{abstract}

%% file: sections/01_introduction.tex
\section{Introduction}

Retrieval-augmented generation (RAG) grounds language models in external
sources of evidence \cite{lewis2020rag}, but multi-hop questions challenge the
standard retrieve-then-generate abstraction. The first useful passage rarely
completes the answer; instead, it changes the information need. It may expose a
bridge entity, reveal a missing relation, indicate that the current query is
misdirected, or provide sufficient support to stop. The next step is therefore
a control decision over an evolving evidence state: the system must decide
whether to continue along the retrieval frontier, reformulate the query,
expand around a bridge entity, request a missing fact, check sufficiency, or
stop and answer.

\begin{figure}[H]
\centering
\includegraphics[width=\columnwidth,height=0.5\textheight,keepaspectratio]{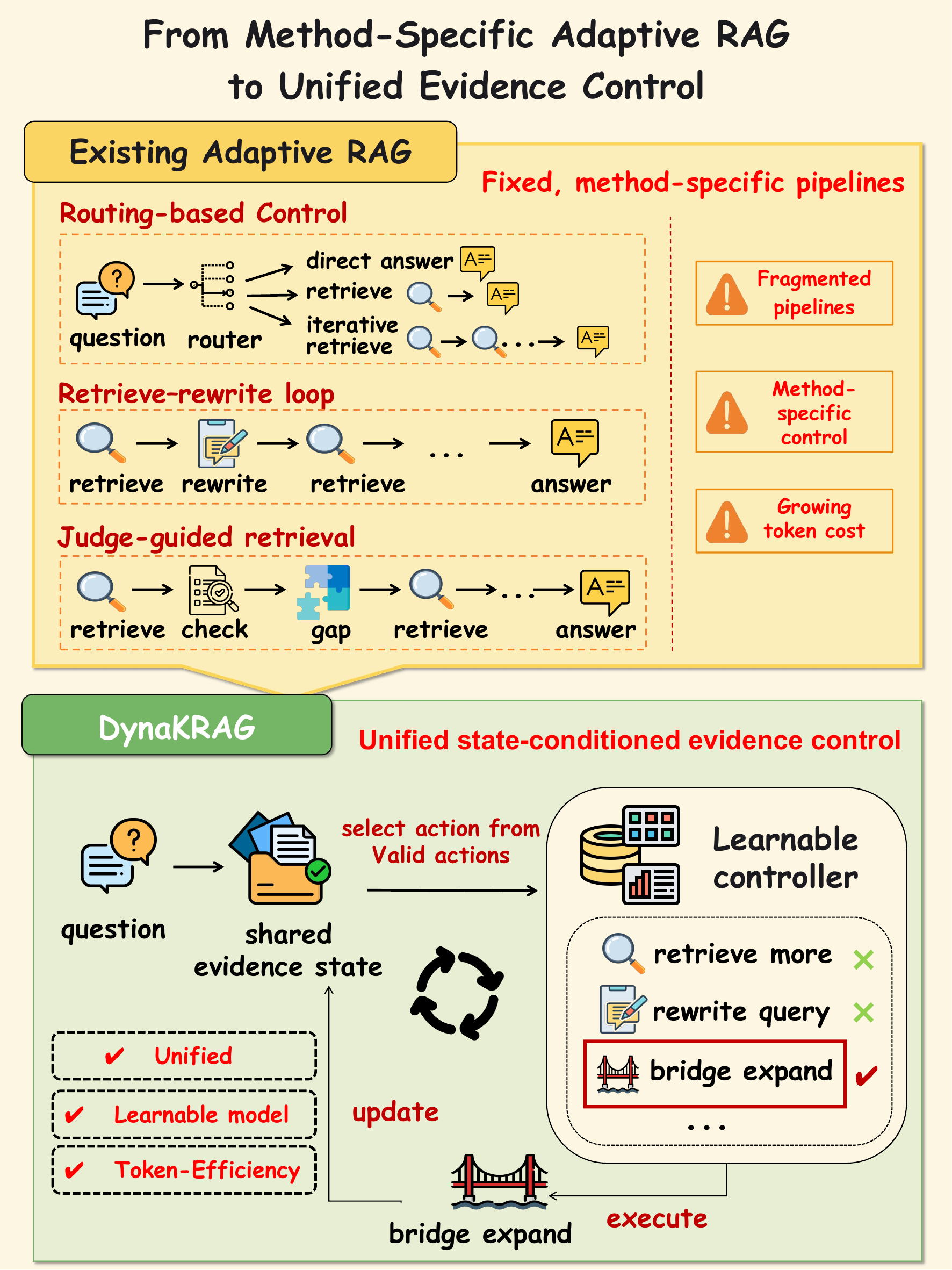}
\caption{Method-specific adaptive RAG pipelines (top) versus the unified
\method controller (bottom), which composes state-valid operations over a
shared evidence state.}
\label{fig:motivation}
\end{figure}

Adaptive RAG methods address parts of this problem by allowing intermediate
results to guide subsequent retrieval and reasoning
\cite{trivedi2023ircot,jeong2024adaptiverag}. Recent systems further introduce
useful behaviors such as query reformulation, evidence critique, sufficiency
checking, and gap-directed retrieval
\cite{jiang2023flare,su2024dragin,asai2023selfrag,yan2024crag,li2026s2grag,li2026par2rag}.
However, these behaviors are still largely embedded in method-specific
pipelines. Each pipeline prescribes its own transition structure, limiting how
different evidence operations can be compared, composed, and learned within a
shared decision process (Figure~\ref{fig:motivation}, top). For example, a
system that can decide whether to retrieve again may not consider query
rewriting or bridge expansion, while a sufficiency module may identify a
missing fact without comparing the available ways to acquire it. These systems
therefore instantiate different solutions to the same underlying problem:
selecting the most appropriate executable evidence operation as the state
evolves.

The choice of operation also determines computational cost. Adaptive RAG
trajectories may repeatedly invoke retrievers and language models, while
different operations consume different amounts of computation. Additional
context can improve evidence coverage, but it can also introduce distractors,
expand downstream prompts, and spend tokens on redundant information. A larger
retrieval budget is therefore not a reliable substitute for better control.
An effective adaptive RAG system should allocate computation to operations
whose expected benefit justifies their cost, avoid invalid or low-value
transitions, and stop once the available evidence is sufficient. This
motivates evaluating the quality--compute operating points exposed by an
evidence-control policy rather than reporting only a single accuracy point. We
accordingly report answer quality together with token use, retrieval calls,
and language-model calls, and examine performance under controlled retrieval
caps.

We introduce \method, a unified learning framework for adaptive evidence
acquisition (Figure~\ref{fig:motivation}, bottom). \method represents heterogeneous RAG behaviors, including frontier retrieval, gap-directed retrieval, query rewriting, bridge-entity expansion, sufficiency checking, and stopping, as atomic operations over a shared evidence state. A deterministic validity layer first removes operations whose prerequisites are unmet or whose
budgets are exhausted. A learned continuation gate then decides whether the
trajectory should stop or continue. Conditional on continuing, a learned
advantage scorer ranks feasible nonterminal operations by their predicted gain
relative to answering from the current state. Executing the selected operation
updates the shared state and may enable different operations at the next step,
turning fixed RAG pipelines into composable sequential decisions.

Training uses two complementary supervision signals. We derive action-level
targets from changes in supporting-evidence recall and train the continuation
gate using the improvement obtained by completing a trajectory rather than
answering from an intermediate prefix. Supporting-evidence annotations and
reference answers are used only to construct training labels; at inference
time, the controller relies exclusively on features of the observable evidence
state. The resulting policy jointly determines whether evidence acquisition
should continue and which currently valid operation should be executed. After
acquisition terminates, an optional terminal compressor constructs a compact,
answer-focused context for the final generator
\cite{xu2024recomp,li2025brief,jeong2025ecorag}. To isolate the contribution of
adaptive evidence control from that of terminal compression, we report both
uncompressed and compressed operating points.

We make three contributions.
\begin{itemize}
\item We formulate adaptive multi-hop evidence acquisition as a unified
evidence-action framework in which heterogeneous RAG strategies are
represented as atomic operations and shared state transitions rather than
isolated, method-specific pipelines.

\item We develop a learned controller that combines deterministic action
validity, learned continuation, and stop-normalized action advantages to
coordinate both whether evidence acquisition should continue and which valid
operation should be executed. The controller is trained using
supporting-evidence and answer-improvement signals while relying only on
observable state features at inference time.

\item We evaluate \method on three multi-hop QA benchmarks and three
answer-generation backbones. \method achieves the best EM and F1 scores across
all nine dataset--backbone settings. Relative to S2G-RAG with matched
backbones, it consistently achieves higher F1 while using fewer total tokens
and retrieval calls. A controller trained on Qwen2.5-7B also transfers directly
to GPT-4o-mini and Llama-3.1-8B without additional policy training, while the
compressed and uncompressed configurations expose complementary
quality--compute operating points.
\end{itemize}

%% file: sections/02_related_work.tex
\section{Related Work}

\paragraph{Multi-hop evidence acquisition.}
Retrieval-augmented generation grounds language-model predictions in external
non-parametric evidence \cite{lewis2020rag,huang2026retrievalsurvey}. Multi-hop benchmarks such as
HotpotQA, 2WikiMultiHopQA, and MuSiQue extend this setting to questions whose
answers depend on evidence distributed across documents and reasoning steps
\cite{yang2018hotpotqa,ho2020twowiki,trivedi2022musique}. As intermediate
facts reveal new entities and relations, the information needed at a later
hop may be unavailable from the original query alone. Multi-hop dense
retrieval addresses this dependency by conditioning later retrieval on
previously acquired evidence \cite{xiong2021mdr}, establishing evidence
acquisition as a trajectory that evolves with the partial solution state.

\paragraph{Iterative and adaptive retrieval.}
Subsequent work develops several mechanisms for steering this trajectory.
ReAct establishes the broader paradigm of interleaving reasoning with
external actions \cite{yao2023react}. IRCoT interleaves retrieval with
generated reasoning, while Self-Ask turns compositional questions into
searchable follow-up questions
\cite{trivedi2023ircot,press2022selfask}.
CoRAG learns multi-step retrieval chains through iterative query
reformulation
\cite{wang2025corag,zhuang2024efficientrag,fang2025kirag},
and RQ-RAG trains models to rewrite, decompose, or disambiguate queries
\cite{chan2024rqrag}. A complementary line
of work adapts retrieval timing and strategy. FLARE and DRAGIN trigger
retrieval from generation-time information needs
\cite{jiang2023flare,su2024dragin,yao2025seakr,baek2025probingrag},
while Adaptive-RAG routes questions
among retrieval regimes according to estimated complexity
\cite{jeong2024adaptiverag,tang2025mbarag,guan2026deeprag}. These methods make retrieval responsive to the
evolving reasoning process, but each controller focuses on a particular
decision, such as constructing the next query, determining retrieval timing,
or selecting a retrieval strategy. Reinforcement learning has also been
explored for multi-turn search and information-seeking agents
\cite{jin2025searchr1,song2025r1searcher,wu2025webdancer,
jiang2025s3,zhao2026rsearch,zheng2025stepsearch,li2025r3rag}. Recent surveys
and benchmarks have begun to systematize this broader space of LLM agents and
agentic information seeking
\cite{zhang2024agenticir,xi2026searchagents,xi2025infodeepseek,zhou2026externalization},
while skill-centric methods synthesize, represent, and retrieve reusable agent
capabilities \cite{wang2026skillsfly,yu2026latentskill,ding2026skill2query}.
In contrast to
these trajectory-level RL methods, \method learns a lightweight one-step
advantage ranker over a constrained action set, without rollouts or
trajectory-level reward optimization.

\paragraph{Evidence diagnosis and action-level control.}
Recent systems increasingly use accumulated evidence to guide subsequent
computation. CRAG evaluates retrieval quality and invokes corrective
processing \cite{yan2024crag,joren2025sufficientcontext}, while Self-RAG learns retrieval and critique
decisions through reflection tokens \cite{asai2023selfrag}. S2G-RAG predicts
evidence sufficiency and converts structured evidence gaps into subsequent
retrieval queries
\cite{li2026s2grag,yang2025simrag,park2025stoprag,chen2026stride};
PAR$^2$-RAG combines breadth-first
evidence coverage with depth-first refinement and sufficiency control
\cite{li2026par2rag,lee2025smr}. Taken together, existing systems support frontier
expansion, query reformulation, gap-directed retrieval, diagnosis, and
stopping. However, these operations are usually studied within separate,
method-specific pipelines, and it remains unclear how a system should choose
among them as its evidence state evolves. \method addresses this decision
problem directly: the current evidence state determines the executable
operations, and a learned advantage scorer selects among them as the
trajectory unfolds.

%% file: sections/03_method.tex
\section{Method}

\subsection{Overview and Problem Formulation}

We consider multi-hop question answering with a question $q$, a corpus
$\mathcal{D}$, a retriever $R$, and an answer generator $G$. Unlike standard
RAG, which commits to a fixed retrieval depth or a prescribed iterative
routine, our setting allows the system to choose a different evidence
operation after each state update. The objective is to acquire sufficient
support for answering $q$ while avoiding invalid, redundant, or unproductive
operations. \method instantiates a unified evidence-action controller without
replacing $R$ or $G$; in our experiments, a controller trained on one answer
backbone is used with other answer backbones without retraining.

Formally, \method constructs an evidence trajectory
$\tau=(s_0,a_0,s_1,a_1,\ldots,a_{T-1},s_T)$. The initial state $s_0$
contains the question and an empty evidence history. At step $t$, the state
$s_t$ summarizes all information observable at inference time, including
retrieved documents, query and action history, retrieval-frontier statistics,
detected bridge entities, and any missing-information feedback. A deterministic
validity layer maps this state to an executable action set
$\mathcal{A}(s_t)$. The learned controller then considers only these valid
actions and selects
$a_t\in\mathcal{A}(s_t)$, while executing $a_t$ produces the next state $s_{t+1}$.
Because a transition can reveal a new gap, entity, or sufficiency judgment,
it can also change which actions are available at the next step. 

Acquisition terminates either when the controller selects \texttt{stop\_answer} or when an
executed \texttt{sufficiency\_check} confirms that the accumulated evidence is
sufficient. Otherwise, the controller continues until the step cap. The
budget-sensitivity experiments additionally impose an explicit retrieval-call
cap. The accumulated evidence is optionally
compressed into an answer-focused context and passed to $G$. Under this
formulation, query construction, operation selection, and retrieval depth
become trajectory-level decisions rather than fixed hyperparameters. The
remainder of this section defines the evidence-action framework, explains how
the state-conditioned policy is learned, and describes terminal compression
and inference.

\begin{figure*}[t]
\centering
\includegraphics[width=0.8\textwidth]{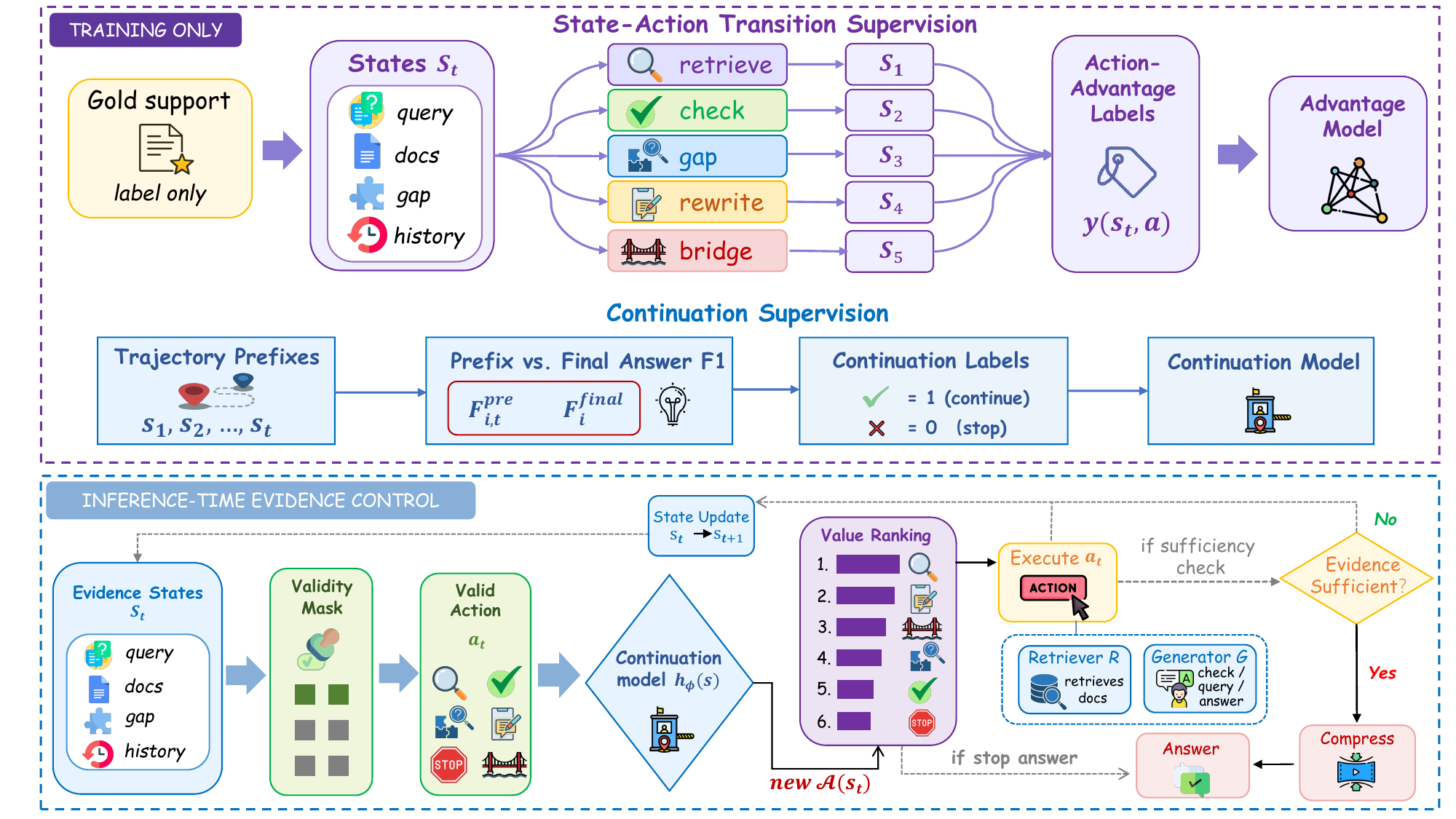}
\caption{Overview of \method training and inference. The framework learns an advantage model and a continuation model for hard gating. Gated actions are ranked and executed until stopping, positive sufficiency, or the control horizon.}
\label{fig:method}
\end{figure*}

\subsection{Unified Evidence-Action Framework}

The evidence state contains the question, retrieved documents, the frontier cursor in the initial ranking, the latest query, an optional missing-information description, a sufficiency result, action history, and accumulated cost. Both
learned models observe only inference-time quantities derived from this state
(the full feature list appears in the supplementary material). Gold answers,
answer scores, supporting facts, and support recall are never used as runtime
features.

Seven atomic operations are defined, six of which participate in the
evidence-control loop. Four produce new evidence: \texttt{retrieve\_more}
($\mathcal{R}$) advances the existing retrieval frontier;
\texttt{gap\_query} ($\mathcal{G}$) generates a query for an identified
missing fact; \texttt{rewrite\_query} ($\mathcal{W}$) reformulates the current
query using accumulated evidence; and \texttt{bridge\_entity\_expand}
($\mathcal{B}$) retrieves around entities detected in the evidence.
\texttt{sufficiency\_check} ($\mathcal{S}$) asks $G$ whether the accumulated
evidence supports an answer. A positive judgment terminates acquisition,
whereas a negative judgment records the missing information and enables
gap-directed retrieval. \texttt{stop\_answer} ($\mathcal{T}$) allows the
controller to terminate acquisition directly without invoking another
evidence operation.
The seventh operation, \texttt{compress\_answer\_evidence} ($\mathcal{C}$),
is applied once after acquisition terminates in the full configuration and is
recorded in the action trace. We handle this component outside the learned ranking process and analyze it separately from the acquisition policy.

Not every action is meaningful in every state. We construct
$\mathcal{A}(s_t)$ before scoring. Reformulation and sufficiency actions
require retrieved evidence, gap querying requires a known gap, bridge
expansion requires a bridge candidate, and retrieval actions require
remaining retrieval budget. This hard validity layer prevents undefined or redundant transitions. The learning component only needs to decide whether to continue and rank the executable operations.

\subsection{Learning a State-Conditioned Control Policy}

The controller compares valid operations by their advantage over stopping.
Let $U(s)=\operatorname{SR}(s)$ denote train-time support utility, where
$\operatorname{SR}(s)$ is the recall of the gold supporting documents under
the current state, and let $T^{+}(s,a)$ be the effective post-action state.
We define
\begin{equation}
\label{eq:action_target}
u_{\lambda,\beta}(s,a)=U(T^{+}(s,a))
-\lambda\mathbb{I}[a\neq a_{\mathrm{stop}}]U(s)
-\beta c_{\mathrm{exec}}(a),
\end{equation}
where $\lambda$ controls the opportunity cost of continuing and $\beta$
weights execution cost.
For acquisition, $T^{+}$ contains the newly retrieved
evidence; for stopping, $T^{+}(s,a_{\mathrm{stop}})=s$; and for sufficiency
checking, it is the diagnostic state that either supports termination or
exposes a gap for targeted retrieval. Because a sufficiency check adds no
evidence, its effective utility is modeled as $U(s)+\rho(s)(1-U(s))$, where
$\rho(s)$ estimates the value of the gap it may expose; the resulting target
is given in the supplementary material. We train on the advantage over
stopping:
\begin{equation}
\begin{aligned}
A_{\lambda,\beta}(s,a)
&=u_{\lambda,\beta}(s,a)-u_{\lambda,\beta}(s,a_{\mathrm{stop}}),\\
A(s,a_{\mathrm{stop}})&=0.
\end{aligned}
\end{equation}
We fit a regressor $\widehat A_{\theta}(s,a)$ to rank valid actions on this
common scale. Training targets are computed counterfactually. For each logged prefix state, we execute every valid action once and record the resulting state and support recall. This ensures that all valid actions from the same state are evaluated on a consistent scale. The result is a supervised one-step regression target
rather than a trajectory-level Bellman value.

Executing the selected action produces the next evidence state.
\texttt{retrieve\_more} advances the initial ranking, while targeted actions
issue a gap-focused, rewritten, or bridge-entity query to the dense index:
the gap-focused query uses the recorded missing-information description, the
rewritten query reformulates the latest query, and the bridge query expands a
detected bridge entity. Each retrieval-producing action merges the top unseen
documents into the evidence state. A \texttt{sufficiency\_check} either confirms answer readiness and terminates
the trajectory or records a missing-information description that enables
subsequent gap-directed retrieval. After a negative judgment, the updated
diagnostic state conditions the next policy decision.

\paragraph{Continuation model and stop suppression.}

The continuation model determines whether a \texttt{stop\_answer} candidate
should terminate acquisition. Its training labels compare the answer quality at each prefix with the quality achieved after completing the trajectory. The continuation target marks prefixes whose answers are still inadequate but can improve substantially through further acquisition. The premature-stop target identifies cases where stopping early would prevent a successful completion. The gate comprises a
continuation model, an answer-gain regressor, and a premature-stop model that
operate on observable state features. After validity and budget filtering, the
gate is evaluated only when stopping and an affordable nonterminal action are
both available. If the gate votes to continue, \texttt{stop\_answer} is removed
from the candidate set, and if it votes to stop, only \texttt{stop\_answer} is
retained. The advantage scorer then ranks the remaining candidates. The process stops when it reaches the budget limit or completes four steps. The supplementary material provides further details on the targets, features, data balancing, and calibration.

\subsection{Terminal Evidence Compression}

After acquisition terminates, the full \method configuration applies
answer-focused evidence compression. The compressor asks $G$ to extract a
small set of mutually supporting snippets from the accumulated documents, and
the final answer is generated from the compressed state. This terminal
operation is excluded from the action-advantage target in
Equation~\ref{eq:action_target} and evaluated by ablation.

\subsection{Inference Procedure}

\begin{algorithm}[!t]
\caption{\method Inference}
\label{alg:dynakrag}
\begin{algorithmic}[1]
\STATE \textbf{Input:} question $q$, corpus $\mathcal{D}$, retriever $R$,
generator $G$
\STATE \textbf{Output:} answer $y$
\STATE $s \leftarrow \operatorname{InitState}(q)$
\FOR{$t=0,\ldots,T-1$}
  \STATE $\mathcal{A} \leftarrow
    \operatorname{FilterBudget}(\operatorname{ValidActions}(s),s)$
  \IF{$\mathcal{A}\setminus\{\texttt{stop\_answer}\}=\varnothing$}
    \STATE \textbf{break}
  \ENDIF
  \IF{$\texttt{stop\_answer}\in\mathcal{A}$}
    \STATE $\mathcal{A}\leftarrow
      \operatorname{ContinuationGate}(s,\mathcal{A})$
  \ENDIF
  \STATE $a\leftarrow\arg\max_{a'\in\mathcal{A}}
         \widehat A_{\theta}(s,a')$
  \IF{$a=\texttt{stop\_answer}$}
    \STATE \textbf{break}
  \ENDIF
  \STATE $s \leftarrow \operatorname{Execute}(a,s;\mathcal{D},R,G)$
\ENDFOR
\STATE \textit{// Optional terminal evidence compression.}
\IF{$\operatorname{CompressionEnabled}()$}
  \STATE $s \leftarrow \operatorname{Compress}(s;G)$
\ENDIF
\STATE $y \leftarrow \operatorname{Generate}(q,\operatorname{Evidence}(s);G)$
\STATE \textbf{return} $y$
\end{algorithmic}
\end{algorithm}

Algorithm~\ref{alg:dynakrag} summarizes inference. The validity layer first
removes actions that cannot be executed, the continuation gate resolves
whether stopping is admissible, and the advantage scorer ranks the resulting
candidate set.

%% file: sections/04_experiments.tex
\section{Experiments}

We evaluate the full system in terms of answer quality, component
contributions, retrieval-budget sensitivity, and performance across question
structures.

\input{tables/main_results}
\input{tables/ablation}

\subsection{Experimental Setup}

\paragraph{Datasets.}
We evaluate \method on HotpotQA \cite{yang2018hotpotqa}, 2Wiki
\cite{ho2020twowiki}, and MuSiQue \cite{trivedi2022musique} on their complete
official evaluation splits. Together, they cover compositional, comparison,
bridge-comparison, and inference questions with two to four hops. Split sizes
and structures are reported in the appendix.

\paragraph{Metrics.}
We report normalized Exact Match (EM) and token-level answer F1, with F1 as
the primary answer-quality metric. For iterative methods, we additionally
record total token consumption (prompt and completion tokens summed over all
model calls in a run), retrieval calls, and language-model calls to evaluate
efficiency. Supporting-evidence recall is used to construct training targets
and diagnose retrieval behavior. For compression analysis, we separately
report tokens in the final answer-generation prompt. This final-prompt
measure counts only the context inspected by the final answer call, unlike
the end-to-end total-token measure.

\paragraph{Baselines.}
We compare against fixed-$K$ RAG \cite{lewis2020rag} and controlled
implementations of IRCoT \cite{trivedi2023ircot}, S2G-RAG
\cite{li2026s2grag}, CoRAG \cite{wang2025corag}, Adaptive-RAG
\cite{jeong2024adaptiverag}, PAR$^2$-RAG \cite{li2026par2rag}, CRAG
\cite{yan2024crag}, and Self-Ask+Search \cite{press2022selfask}. The baselines
span static, iterative, adaptive, corrective, and decomposition-based
retrieval methods. We report CoRAG with three retrieval steps (CoRAG-s3)
in all dataset--backbone pairs. Fixed-$K$ reports the strongest fixed
retrieval depth among settings that do not use gold evidence.

\paragraph{Implementation details.}
We evaluate Qwen2.5-7B-Instruct \cite{qwen2024qwen25},
GPT-4o-mini, and Llama-3.1-8B-Instruct as answer-model backbones. Within each
backbone, all methods use the same corpus, retrieval resources, and evaluation
pipeline. We train
dataset-specific action-advantage and continuation models on 1,000 training
examples disjoint from each reported evaluation split. Supporting-evidence
annotations supervise action advantages, while prefix-to-final answer improvement
supervises continuation. Questions, rather than individual prefixes, are split
80/20 for fitting and calibration; prefix counts are reported in the
supplementary material. Inference uses only observable state features. Both
models are trained on Qwen2.5-7B-Instruct trajectories and transferred to
GPT-4o-mini and Llama-3.1-8B-Instruct without additional policy training.
Each random forest contains 300 estimators and uses a minimum leaf size of 8.
For each dataset, we independently train the controller models three times
and report the mean performance across the resulting runs.

Inference uses deterministic decoding. Main runs add up to three documents
per retrieval operation, allow at most four acquisition or control steps, and
retain at most 12 documents in the final context. Targeted actions query a
BGE-large-en-v1.5 dense index \cite{xiao2023cpack} and retain unseen
documents in score order. The full system applies terminal evidence
compression before answer generation.

\subsection{Main Results}

Table~\ref{tab:main_results} establishes a consistent accuracy lead: \method
ranks first among the compared systems in both EM and F1 in every
dataset--backbone pair (9/9 for each metric), so the gains are not limited to
token-overlap F1. Qwen F1 reaches 0.5998, 0.5340, and 0.3061 on HotpotQA,
2Wiki, and MuSiQue---gains of 2.88, 7.19, and 0.62 points over the strongest
controlled baseline. GPT-4o-mini F1 reaches 0.6218, 0.6391, and 0.3977,
with gains of 1.10, 1.33, and 1.01 points. The strongest baseline varies
across settings (S2G-RAG, CoRAG-s3, CRAG, IRCoT, or Self-Ask+Search), so
these deltas differ from the S2G-RAG comparison below.

The Qwen-trained policy also transfers without modification to
Llama-3.1-8B-Instruct, with no target-backbone retraining. Its F1 scores
of 0.4876, 0.3933, and 0.2391 lead the strongest same-backbone baselines by
1.06, 2.03, and 2.65 points. Ranking first in all six non-Qwen settings
demonstrates effective transfer to both backbones beyond the one used to
collect the controller's training trajectories.

The largest gain occurs on 2Wiki compositional questions, where \method
improves over fixed-$K$ by 24.31 F1 points; the analyses below examine this
structure sensitivity and the contributions of individual components.

\paragraph{Quality and efficiency relative to S2G-RAG.}
S2G-RAG is a strong iterative comparator whose total-token accounting matches
\method's. In all nine matched pairs, \method has higher F1, 10.1--34.3\%
fewer total tokens, and 15.1--43.4\% fewer retrieval calls, thus
Pareto-dominating S2G-RAG under these measures. The execution-cost setting is
reported in the appendix.

\paragraph{Complementary operating points.}
The uncompressed Qwen variant reaches F1 scores of 0.5826, 0.4963, and 0.3008
with 31.5--38.8\% fewer tokens than S2G-RAG. It exceeds S2G-RAG on all three
datasets. Terminal compression moves
to a higher-quality point: F1 rises by 1.72, 3.77, and 0.53 points while the
final-answer prompt shrinks by 54.4\%, 66.3\%, and 71.5\%.
Full statistics appear in the supplement.

\subsection{Ablation Studies}

Table~\ref{tab:ablation} factorizes learned advantage ranking and learned
continuation in a $2{\times}2$ design. Uniform-valid action selection
replaces learned advantage ranking; fixed horizon disables learned continuation
and disallows early \texttt{stop\_answer} while a valid nonterminal action
remains. All four cells hold the validity layer, action set, prompts, budgets,
and terminal compression fixed. The full model performs best on all three
datasets. Replacing learned advantage ranking with uniform-valid selection
while retaining learned continuation reduces F1 by 2.76, 2.13, and 1.24 points
on HotpotQA, 2Wiki, and MuSiQue, respectively. Disabling learned continuation
while retaining advantage ranking produces smaller drops of 0.79, 0.22, and
0.69 points.

The remaining rows retain both learned mechanisms. Removing sufficiency
checking lowers F1 by 1.63, 7.17, and 2.61 points, while removing terminal
compression lowers it by 1.72, 3.77, and 0.53 points. The sufficiency gap is a
combined trajectory-level effect, since removing the probe also disables
gap-directed queries. Removing compression defines the lower-cost operating
point above.

\subsection{Diagnostic Analyses}

\begin{figure}[t]
\centering
\includegraphics[width=\columnwidth]{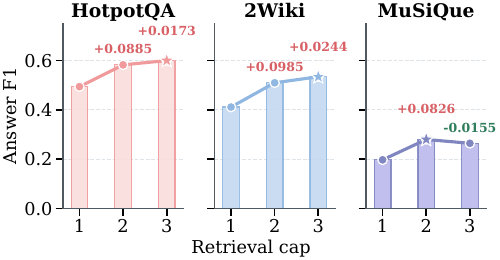}
\caption{F1 under retrieval-call caps; stars mark best cap.}
\label{fig:budget}
\end{figure}

\paragraph{Retrieval-budget sensitivity.}
Figure~\ref{fig:budget} reports F1 under controlled retrieval-call caps.
HotpotQA and 2Wiki improve from one to three calls. The main runs cap control
steps at four rather than retrieval calls, so an example may issue a fourth
retrieval. Within a given budget, \method can allocate calls among
frontier expansion, gap queries, and bridge expansion as the evidence state
changes, rather than repeating a single retrieval operation.

\begin{figure}[t]
\centering
\includegraphics[width=\columnwidth]{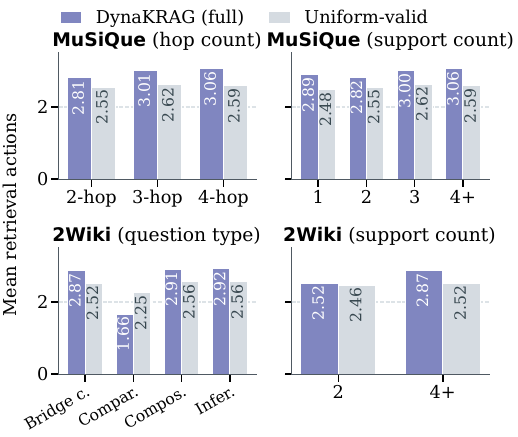}
\caption{Mean retrieval actions
($\mathcal{R}{+}\mathcal{G}{+}\mathcal{W}{+}\mathcal{B}$) by question
structure. \method scales with MuSiQue hops and 2Wiki type; uniform-valid
remains nearly flat. The 2Wiki 4+ bucket is bridge-comparison.}
\label{fig:retrieval_allocation}
\end{figure}

\paragraph{Adaptive retrieval allocation.}
The learned controller adapts how much retrieval it spends to the structure of
the question (Figure~\ref{fig:retrieval_allocation}). Retrieval rises
monotonically from MuSiQue two-hop to four-hop questions (2.81 to 3.06 calls),
while on 2Wiki, comparison questions receive only 1.66 retrieval actions versus
about 2.9 for the remaining types. The uniform-valid comparator allocates
nearly the same budget in every bucket (2.25--2.62), so this variation is
learned rather than inherited from the validity layer. Support-count buckets,
a coarser structure axis available on both datasets, show the same pattern.

\begin{figure}[t]
\centering
\includegraphics[width=0.9\columnwidth]{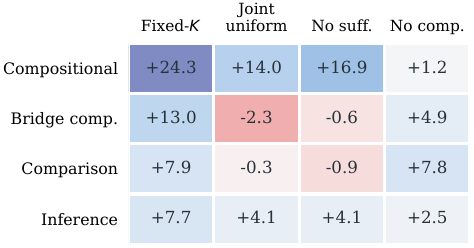}
\caption{Full-controller $\Delta$F1 gains over four comparators on 2Wiki, by question type. Joint uniform disables learned ranking and continuation; No suff./comp. remove the respective components.}
\label{fig:question_structure}
\end{figure}

\paragraph{Question-structure results.}
Figure~\ref{fig:question_structure} breaks down performance on the full 2Wiki
development set, which has 2,751 bridge-comparison, 3,040 comparison, 5,236
compositional, and 1,549 inference questions. Relative to fixed-$K$, \method
gains 12.97, 7.95, 24.31, and 7.68 F1 points on these four groups,
respectively. The largest gain occurs on the compositional group, where
several facts must be assembled across steps. Because Joint uniform disables both
learned components jointly, it does not isolate their individual contributions
and is used only for descriptive structure analysis; it does not replace the
factorial ablation in Table~\ref{tab:ablation}.

On MuSiQue, the full method obtains F1 scores of 0.3694, 0.2691, and 0.1798 on
the 1,252 two-hop, 760 three-hop, and 405 four-hop questions, respectively.
The corresponding S2G-RAG scores are 0.3568, 0.2562, and 0.2060.

%% file: tables/main_results.tex
\begin{table*}[t]
\centering
\small
\setlength{\tabcolsep}{2.2pt}
\renewcommand{\arraystretch}{1.04}
\begin{tabular*}{\textwidth}{@{\extracolsep{\fill}}ll*{9}{r}@{}}
\toprule
\multicolumn{1}{c}{\multirow{2}{*}{\textbf{Backbone}}} &
\multicolumn{1}{c}{\multirow{2}{*}{\textbf{Method}}} &
\multicolumn{3}{c}{\textbf{HotpotQA}} &
\multicolumn{3}{c}{\textbf{2Wiki}} &
\multicolumn{3}{c}{\textbf{MuSiQue}} \\
\cmidrule(lr){3-5}\cmidrule(lr){6-8}\cmidrule(lr){9-11}
 & & \textbf{EM$\uparrow$} & \textbf{F1$\uparrow$} & \textbf{Tok.$\downarrow$} &
\textbf{EM$\uparrow$} & \textbf{F1$\uparrow$} & \textbf{Tok.$\downarrow$} &
\textbf{EM$\uparrow$} & \textbf{F1$\uparrow$} & \textbf{Tok.$\downarrow$} \\
\midrule
\multirow{9}{*}{\texttt{Qwen2.5-7B}} & Fixed-K & 0.3958 & 0.5177 & 1892.3 & 0.3177 & 0.3757 & 3088.9 & 0.1518 & 0.2553 & 2918.3 \\
 & IRCoT & 0.4339 & 0.5617 & 1723.5 & 0.3514 & 0.4308 & 2441.8 & 0.1862 & 0.2889 & \best{2109.6} \\
 & S2G-RAG & \second{0.4425} & \second{0.5710} & 2807.3 & \second{0.3818} & \second{0.4621} & 3543.5 & 0.1936 & \second{0.2999} & 3886.8 \\
 & CoRAG-s3 & 0.4282 & 0.5573 & 2814.3 & 0.3490 & 0.4273 & 3675.8 & 0.1858 & 0.2848 & 3263.6 \\
 & Adaptive-RAG & 0.3862 & 0.5042 & \best{1596.4} & 0.2976 & 0.3557 & \best{2180.0} & 0.1647 & 0.2579 & \second{2160.4} \\
 & PAR$^2$-RAG & 0.3779 & 0.4952 & 3053.6 & 0.3071 & 0.3691 & 3971.2 & 0.1448 & 0.2480 & 3755.6 \\
 & CRAG & 0.4181 & 0.5445 & \second{1720.1} & 0.3437 & 0.4099 & \second{2375.4} & 0.1676 & 0.2740 & 2479.5 \\
 & Self-Ask+Search & 0.4043 & 0.5273 & 3302.0 & 0.3494 & 0.4345 & 4565.5 & \second{0.1974} & 0.2998 & 4761.8 \\
\fullrowshade{ourblue} & \method & \best{0.4732} & \best{0.5998} & 2373.1 & \best{0.4587} & \best{0.5340} & 3077.9 & \best{0.1986} & \best{0.3061} & 3082.3 \\
\midrule
\multirow{9}{*}{\texttt{GPT-4o-mini}} & Fixed-K & 0.4170 & 0.5781 & 1745.1 & 0.4152 & 0.5204 & 3097.8 & 0.1899 & 0.3279 & 2763.6 \\
 & IRCoT & 0.4485 & 0.6083 & 1545.7 & 0.4878 & 0.6188 & 2217.1 & 0.2453 & 0.3870 & \second{1939.6} \\
 & S2G-RAG & \second{0.4498} & 0.6095 & 1766.0 & \second{0.4972} & \second{0.6258} & 2641.3 & 0.2280 & 0.3691 & 3091.2 \\
 & CoRAG-s3 & 0.4496 & \second{0.6108} & 2569.7 & 0.4912 & 0.6173 & 3286.8 & \second{0.2482} & 0.3872 & 3046.0 \\
 & Adaptive-RAG & 0.3419 & 0.4783 & \best{1120.4} & 0.3636 & 0.4311 & \best{1504.6} & 0.1928 & 0.3172 & \best{1789.1} \\
 & PAR$^2$-RAG & 0.3918 & 0.5502 & 2776.0 & 0.4173 & 0.5354 & 3536.9 & 0.1887 & 0.3241 & 3477.4 \\
 & CRAG & 0.4406 & 0.5951 & \second{1355.4} & 0.4734 & 0.5856 & \second{1830.5} & 0.2151 & 0.3578 & 2020.7 \\
 & Self-Ask+Search & 0.4404 & 0.6023 & 3432.1 & 0.4740 & 0.6075 & 4525.7 & 0.2449 & \second{0.3876} & 4570.0 \\
\fullrowshade{ourblue} & \method & \best{0.4922} & \best{0.6218} & 1588.5 & \best{0.5545} & \best{0.6391} & 2169.0 & \best{0.2635} & \best{0.3977} & 2499.9 \\
\midrule
\multirow{9}{*}{\texttt{Llama-3.1-8B}} & Fixed-K & 0.3029 & 0.4311 & 1799.4 & 0.1489 & 0.2444 & 3025.8 & 0.0695 & 0.1461 & 2814.2 \\
 & IRCoT & 0.3252 & 0.4466 & 1657.1 & 0.2901 & \second{0.3730} & 2249.4 & \second{0.1316} & \second{0.2126} & 2049.1 \\
 & S2G-RAG & 0.3517 & 0.4761 & 2843.0 & 0.2976 & 0.3725 & 3369.2 & 0.1146 & 0.1973 & 3848.5 \\
 & CoRAG-s3 & 0.3284 & 0.4499 & 2576.3 & 0.2577 & 0.3341 & 3135.5 & 0.0956 & 0.1690 & 2945.1 \\
 & Adaptive-RAG & 0.3215 & 0.4395 & \second{1197.2} & 0.2476 & 0.3116 & \second{1706.6} & 0.0952 & 0.1719 & \second{1749.8} \\
 & PAR$^2$-RAG & 0.3156 & 0.4321 & 3035.5 & 0.2499 & 0.3168 & 3551.4 & 0.0881 & 0.1674 & 3333.6 \\
 & CRAG & \second{0.3564} & \second{0.4770} & \best{968.8} & \second{0.3087} & 0.3668 & \best{1192.2} & 0.1076 & 0.1980 & \best{1477.9} \\
 & Self-Ask+Search & 0.3211 & 0.4354 & 3525.1 & 0.2774 & 0.3552 & 4454.7 & 0.0935 & 0.1685 & 4000.1 \\
\fullrowshade{ourblue} & \method & \best{0.3742} & \best{0.4876} & 1868.6 & \best{0.3201} & \best{0.3933} & 2915.0 & \best{0.1514} & \best{0.2391} & 3254.6 \\
\bottomrule
\end{tabular*}
\caption{Accuracy and token use on the full evaluation splits. The
Qwen-trained controller transfers to the other backbones. Tok.\ is mean token
use ($\downarrow$); Fixed-$K$ reports prompt rather than total tokens and is
excluded from token rankings. \textbf{Bold}/\underline{underline}: best/second
best per backbone.}
\label{tab:main_results}
\end{table*}

%% file: tables/ablation.tex
\begin{table}[t]
\centering
\footnotesize
\setlength{\tabcolsep}{2.4pt}
\renewcommand{\arraystretch}{0.96}
\begin{tabularx}{\columnwidth}{>{\raggedright\arraybackslash}X*{5}{>{\centering\arraybackslash}c}} \toprule \textbf{Variant} & \shortstack{\textbf{Learned}\\\textbf{adv. rank}} & \shortstack{\textbf{Learned}\\\textbf{cont.}} & \textbf{HQA} & \textbf{2Wiki} & \textbf{MuS} \\ \midrule \rowcolor{ourblue} \method (full) & $\checkmark$ & $\checkmark$ & \textbf{0.5998} & \textbf{0.5340} & \textbf{0.3061} \\ Uniform-valid & $\times$ & $\checkmark$ & 0.5722 & 0.5127 & 0.2937 \\ Fixed horizon & $\checkmark$ & $\times$ & \second{0.5919} & \second{0.5318} & 0.2992 \\ Uniform + fixed & $\times$ & $\times$ & 0.5762 & 0.5156 & 0.2957 \\ \midrule No sufficiency & $\checkmark$ & $\checkmark$ & 0.5835 & 0.4623 & 0.2800 \\ No compression & $\checkmark$ & $\checkmark$ & 0.5826 & 0.4963 & \second{0.3008} \\ \bottomrule \end{tabularx}
\caption{Qwen2.5-7B factorized F1 ablations. $\checkmark/\times$ denote
learned/uniform advantage ranking and learned/fixed-horizon continuation; other
components are fixed. The last two rows remove the indicated component from
the full model. \textbf{Bold}/\underline{underline}: best/second best per
dataset.}
\label{tab:ablation}
\end{table}

%% file: sections/07_conclusion.tex
\section{Conclusion}

We introduced \method, a controller that decides how to gather and use
evidence for multi-hop RAG. Instead of following a fixed pipeline, it chooses
valid actions as the evidence changes and decides when to answer. Across three
datasets and three answer models, \method achieves the best EM and F1 in all
nine settings. It also outperforms the baseline method while using fewer
tokens and retrieval calls. Terminal compression further improves answer
quality by giving the answer model a shorter, more focused context. Our
analyses show that better results come from choosing useful actions and
stopping at the right time, rather than simply retrieving more documents.

%% file: main.bbl
\begin{thebibliography}{48}
\providecommand{\natexlab}[1]{#1}

\bibitem[{Asai et~al.(2024)Asai, Wu, Wang, Sil, and
  Hajishirzi}]{asai2023selfrag}
Asai, A.; Wu, Z.; Wang, Y.; Sil, A.; and Hajishirzi, H. 2024.
\newblock {Self-RAG}: Learning to Retrieve, Generate, and Critique through
  Self-Reflection.
\newblock In \emph{International Conference on Learning Representations}.

\bibitem[{Baek et~al.(2025)Baek, Chang, Kim, Lee, and Lee}]{baek2025probingrag}
Baek, I.; Chang, H.; Kim, B.; Lee, J.; and Lee, H. 2025.
\newblock {Probing-RAG}: Self-Probing to Guide Language Models in Selective
  Document Retrieval.
\newblock In \emph{Findings of the Association for Computational Linguistics:
  NAACL 2025}, 3287--3304. Albuquerque, New Mexico: Association for
  Computational Linguistics.

\bibitem[{Chan et~al.(2024)Chan, Xu, Yuan, Luo, Xue, Guo, and
  Fu}]{chan2024rqrag}
Chan, C.-M.; Xu, C.; Yuan, R.; Luo, H.; Xue, W.; Guo, Y.; and Fu, J. 2024.
\newblock {RQ-RAG}: Learning to Refine Queries for Retrieval Augmented
  Generation.
\newblock In \emph{Conference on Language Modeling}.

\bibitem[{Chen et~al.(2026)Chen, Zhao, Zheng, Hou, and Xu}]{chen2026stride}
Chen, W.; Zhao, L.; Zheng, Z.; Hou, H.; and Xu, T. 2026.
\newblock {STRIDE}: Strategic Iterative Decision-Making for Retrieval-Augmented
  Multi-Hop Question Answering.
\newblock In \emph{Proceedings of the 49th International ACM SIGIR Conference
  on Research and Development in Information Retrieval}, 202--212. ACM.

\bibitem[{Ding et~al.(2026)Ding, Guo, Lu, Zhou, Zhou, Zhang, Lin, and
  Ge}]{ding2026skill2query}
Ding, L.; Guo, Z.; Lu, B.; Zhou, C.; Zhou, Y.; Zhang, W.; Lin, J.; and Ge, D.
  2026.
\newblock {Skill2Query}: Exploiting Skill Structure to Generate Pseudo-Queries
  for Agent Skill Retrieval.
\newblock \emph{arXiv preprint arXiv:2608.16071}.

\bibitem[{Fang, Meng, and MacDonald(2025)}]{fang2025kirag}
Fang, J.; Meng, Z.; and MacDonald, C. 2025.
\newblock {KiRAG}: Knowledge-Driven Iterative Retriever for Enhancing
  Retrieval-Augmented Generation.
\newblock In \emph{Proceedings of the 63rd Annual Meeting of the Association
  for Computational Linguistics (Volume 1: Long Papers)}, 18969--18985. Vienna,
  Austria: Association for Computational Linguistics.

\bibitem[{Guan et~al.(2026)Guan, Zeng, Meng, Xin, Lu, Lin, Han, Sun, and
  Zhou}]{guan2026deeprag}
Guan, X.; Zeng, J.; Meng, F.; Xin, C.; Lu, Y.; Lin, H.; Han, X.; Sun, L.; and
  Zhou, J. 2026.
\newblock {DeepRAG}: Thinking to Retrieve Step by Step for Large Language
  Models.
\newblock In \emph{International Conference on Learning Representations},
  volume 2026, 117056--117071.

\bibitem[{Ho et~al.(2020)Ho, Nguyen, Sugawara, and Aizawa}]{ho2020twowiki}
Ho, X.; Nguyen, A.-K.~D.; Sugawara, S.; and Aizawa, A. 2020.
\newblock Constructing A Multi-Hop {QA} Dataset for Comprehensive Evaluation of
  Reasoning Steps.
\newblock In \emph{Proceedings of the 28th International Conference on
  Computational Linguistics}, 6609--6625.

\bibitem[{Huang et~al.(2026)Huang, Chen, Lin, Qin, Feng, Zhang, and
  Yu}]{huang2026retrievalsurvey}
Huang, J.; Chen, J.; Lin, J.; Qin, J.; Feng, Z.; Zhang, W.; and Yu, Y. 2026.
\newblock A Comprehensive Survey on Retrieval Methods in Recommender Systems.
\newblock \emph{ACM Transactions on Information Systems}, 44(1): 1--43.

\bibitem[{Jeong et~al.(2024)Jeong, Baek, Cho, Hwang, and
  Park}]{jeong2024adaptiverag}
Jeong, S.; Baek, J.; Cho, S.; Hwang, S.~J.; and Park, J. 2024.
\newblock {Adaptive-RAG}: Learning to Adapt Retrieval-Augmented Large Language
  Models through Question Complexity.
\newblock In \emph{Proceedings of the 2024 Conference of the North American
  Chapter of the Association for Computational Linguistics: Human Language
  Technologies}, 7036--7050.

\bibitem[{Jeong et~al.(2025)Jeong, Kim, Lee, and Hwang}]{jeong2025ecorag}
Jeong, Y.; Kim, J.; Lee, D.; and Hwang, S.-w. 2025.
\newblock {EC}o{RAG}: Evidentiality-guided Compression for Long Context {RAG}.
\newblock In \emph{Findings of the Association for Computational Linguistics:
  ACL 2025}, 26607--26628. Vienna, Austria: Association for Computational
  Linguistics.

\bibitem[{Jiang et~al.(2025)Jiang, Xu, Lin, Xiao, Wang, Sun, and
  Han}]{jiang2025s3}
Jiang, P.; Xu, X.; Lin, J.; Xiao, J.; Wang, Z.; Sun, J.; and Han, J. 2025.
\newblock s3: You Don{'}t Need That Much Data to Train a Search Agent via {RL}.
\newblock In \emph{Proceedings of the 2025 Conference on Empirical Methods in
  Natural Language Processing}, 21599--21617. Suzhou, China: Association for
  Computational Linguistics.

\bibitem[{Jiang et~al.(2023)Jiang, Xu, Gao, Sun, Liu, Dwivedi-Yu, Yang, Callan,
  and Neubig}]{jiang2023flare}
Jiang, Z.; Xu, F.; Gao, L.; Sun, Z.; Liu, Q.; Dwivedi-Yu, J.; Yang, Y.; Callan,
  J.; and Neubig, G. 2023.
\newblock Active Retrieval Augmented Generation.
\newblock In \emph{Proceedings of the 2023 Conference on Empirical Methods in
  Natural Language Processing}, 7969--7992.

\bibitem[{Jin et~al.(2025)Jin, Zeng, Yue, Yoon, Arik, Wang, Zamani, and
  Han}]{jin2025searchr1}
Jin, B.; Zeng, H.; Yue, Z.; Yoon, J.; Arik, S.~O.; Wang, D.; Zamani, H.; and
  Han, J. 2025.
\newblock {Search-R1}: Training {LLM}s to Reason and Leverage Search Engines
  with Reinforcement Learning.
\newblock In \emph{Second Conference on Language Modeling}.

\bibitem[{Joren et~al.(2025)Joren, Zhang, Ferng, Juan, Taly, and
  Rashtchian}]{joren2025sufficientcontext}
Joren, H.; Zhang, J.; Ferng, C.-S.; Juan, D.-C.; Taly, A.; and Rashtchian, C.
  2025.
\newblock Sufficient Context: A New Lens on Retrieval Augmented Generation
  Systems.
\newblock In \emph{International Conference on Learning Representations},
  volume 2025, 20310--20334.

\bibitem[{Lee, Jeong, and Hwang(2025)}]{lee2025smr}
Lee, D.; Jeong, Y.; and Hwang, S.-w. 2025.
\newblock From Token to Action: State Machine Reasoning to Mitigate
  Overthinking in Information Retrieval.
\newblock In \emph{Findings of the Association for Computational Linguistics:
  EMNLP 2025}, 7048--7064. Suzhou, China: Association for Computational
  Linguistics.

\bibitem[{Lewis et~al.(2020)Lewis, Perez, Piktus, Petroni, Karpukhin, Goyal,
  Kuttler, Lewis, Yih, Rockt{\"a}schel, Riedel, and Kiela}]{lewis2020rag}
Lewis, P.; Perez, E.; Piktus, A.; Petroni, F.; Karpukhin, V.; Goyal, N.;
  Kuttler, H.; Lewis, M.; Yih, W.-t.; Rockt{\"a}schel, T.; Riedel, S.; and
  Kiela, D. 2020.
\newblock Retrieval-Augmented Generation for Knowledge-Intensive {NLP} Tasks.
\newblock In \emph{Advances in Neural Information Processing Systems}.
\newblock ArXiv:2005.11401.

\bibitem[{Li et~al.(2026{\natexlab{a}})Li, Zou, Lv, Zhang, and
  Zhou}]{li2026s2grag}
Li, M.; Zou, J.; Lv, X.; Zhang, C.; and Zhou, G. 2026{\natexlab{a}}.
\newblock {S2G-RAG}: Structured Sufficiency and Gap Judging for Iterative
  Retrieval-Augmented {QA}.
\newblock ArXiv:2604.23783.

\bibitem[{Li et~al.(2026{\natexlab{b}})Li, Wang, Wang, Guo, Li, Sheng, Ravi,
  and Roth}]{li2026par2rag}
Li, X.; Wang, R.; Wang, Y.; Guo, M.; Li, C.; Sheng, T.; Ravi, S.; and Roth, D.
  2026{\natexlab{b}}.
\newblock {PAR$^2$-RAG}: Planned Active Retrieval and Reasoning for Multi-Hop
  Question Answering.
\newblock ArXiv:2603.29085.

\bibitem[{Li et~al.(2025{\natexlab{a}})Li, Gu, Wu, Chang, and
  Peng}]{li2025brief}
Li, Y.; Gu, J.-C.; Wu, D.; Chang, K.-W.; and Peng, N. 2025{\natexlab{a}}.
\newblock {BRIEF}: Bridging Retrieval and Inference for Multi-hop Reasoning via
  Compression.
\newblock In \emph{Findings of the Association for Computational Linguistics:
  NAACL 2025}, 5464--5485. Albuquerque, New Mexico: Association for
  Computational Linguistics.

\bibitem[{Li et~al.(2025{\natexlab{b}})Li, Luo, Li, Li, Cheng, Wang, Zheng,
  Wang, Yin, and Qiu}]{li2025r3rag}
Li, Y.; Luo, Q.; Li, X.; Li, B.; Cheng, Q.; Wang, B.; Zheng, Y.; Wang, Y.; Yin,
  Z.; and Qiu, X. 2025{\natexlab{b}}.
\newblock {R3-RAG}: Learning Step-by-Step Reasoning and Retrieval for {LLM}s
  via Reinforcement Learning.
\newblock In \emph{Findings of the Association for Computational Linguistics:
  EMNLP 2025}, 10491--10507. Suzhou, China: Association for Computational
  Linguistics.

\bibitem[{Park, Cho, and Lee(2025)}]{park2025stoprag}
Park, J.; Cho, S.; and Lee, J.-Y. 2025.
\newblock {Stop-RAG}: Value-Based Retrieval Control for Iterative {RAG}.
\newblock NeurIPS 2025 Workshop on Multi-Turn Interactions in Large Language
  Models; arXiv:2510.14337, arXiv:2510.14337.

\bibitem[{Press et~al.(2023)Press, Zhang, Min, Schmidt, Smith, and
  Lewis}]{press2022selfask}
Press, O.; Zhang, M.; Min, S.; Schmidt, L.; Smith, N.~A.; and Lewis, M. 2023.
\newblock Measuring and Narrowing the Compositionality Gap in Language Models.
\newblock In \emph{Findings of the Association for Computational Linguistics:
  EMNLP 2023}, 5687--5711.

\bibitem[{{Qwen Team} et~al.(2024){Qwen Team}, Yang, Yang, Zhang, Hui, Zheng,
  Yu, Li, Liu, Huang, Wei et~al.}]{qwen2024qwen25}
{Qwen Team}; Yang, A.; Yang, B.; Zhang, B.; Hui, B.; Zheng, B.; Yu, B.; Li, C.;
  Liu, D.; Huang, F.; Wei, H.; et~al. 2024.
\newblock {Qwen2.5} Technical Report.
\newblock ArXiv:2412.15115.

\bibitem[{Song et~al.(2025)Song, Jiang, Min, Chen, Chen, Zhao, Fang, and
  Wen}]{song2025r1searcher}
Song, H.; Jiang, J.; Min, Y.; Chen, J.; Chen, Z.; Zhao, W.~X.; Fang, L.; and
  Wen, J.-R. 2025.
\newblock {R1-Searcher}: Incentivizing the Search Capability in {LLM}s via
  Reinforcement Learning.
\newblock \emph{arXiv preprint arXiv:2503.05592}.

\bibitem[{Su et~al.(2024)Su, Tang, Ai, Wu, and Liu}]{su2024dragin}
Su, W.; Tang, Y.; Ai, Q.; Wu, Z.; and Liu, Y. 2024.
\newblock {DRAGIN}: Dynamic Retrieval Augmented Generation based on the
  Real-time Information Needs of Large Language Models.
\newblock In \emph{Proceedings of the 62nd Annual Meeting of the Association
  for Computational Linguistics (Volume 1: Long Papers)}, 12991--13013.

\bibitem[{Tang et~al.(2025)Tang, Gao, Li, Du, Li, and Xie}]{tang2025mbarag}
Tang, X.; Gao, Q.; Li, J.; Du, N.; Li, Q.; and Xie, S. 2025.
\newblock {MBA-RAG}: a Bandit Approach for Adaptive Retrieval-Augmented
  Generation through Question Complexity.
\newblock In \emph{Proceedings of the 31st International Conference on
  Computational Linguistics}, 3248--3254. Abu Dhabi, UAE: Association for
  Computational Linguistics.

\bibitem[{Trivedi et~al.(2022)Trivedi, Balasubramanian, Khot, and
  Sabharwal}]{trivedi2022musique}
Trivedi, H.; Balasubramanian, N.; Khot, T.; and Sabharwal, A. 2022.
\newblock {MuSiQue}: Multihop Questions via Single-Hop Question Composition.
\newblock \emph{Transactions of the Association for Computational Linguistics},
  10: 539--554.

\bibitem[{Trivedi et~al.(2023)Trivedi, Balasubramanian, Khot, and
  Sabharwal}]{trivedi2023ircot}
Trivedi, H.; Balasubramanian, N.; Khot, T.; and Sabharwal, A. 2023.
\newblock Interleaving Retrieval with Chain-of-Thought Reasoning for
  Knowledge-Intensive Multi-Step Questions.
\newblock In \emph{Proceedings of the 61st Annual Meeting of the Association
  for Computational Linguistics}, 10014--10037.

\bibitem[{Wang et~al.(2026)Wang, Zhou, Fu, Wang, Liu, Zhang, and
  Lin}]{wang2026skillsfly}
Wang, J.; Zhou, C.; Fu, Z.; Wang, J.; Liu, W.; Zhang, W.; and Lin, J. 2026.
\newblock Skills on the Fly: Test-Time Adaptive Skill Synthesis for {LLM}
  Agents.
\newblock \emph{arXiv preprint arXiv:2605.16986}.

\bibitem[{Wang et~al.(2025)Wang, Chen, Yang, Huang, Dou, and
  Wei}]{wang2025corag}
Wang, L.; Chen, H.; Yang, N.; Huang, X.; Dou, Z.; and Wei, F. 2025.
\newblock Chain-of-Retrieval Augmented Generation.
\newblock In \emph{Advances in Neural Information Processing Systems}.

\bibitem[{Wu et~al.(2025)Wu, Li, Fang, Yin, Zhang, Wang, Tao, Zhang, Xi, Tang,
  Jiang, Xie, Huang, and Zhou}]{wu2025webdancer}
Wu, J.; Li, B.; Fang, R.; Yin, W.; Zhang, L.; Wang, Z.; Tao, Z.; Zhang, D.-C.;
  Xi, Z.; Tang, R.; Jiang, Y.; Xie, P.; Huang, F.; and Zhou, J. 2025.
\newblock {WebDancer}: Towards Autonomous Information Seeking Agency.
\newblock In Belgrave, D.; Zhang, C.; Lin, H.; Pascanu, R.; Koniusz, P.;
  Ghassemi, M.; and Chen, N., eds., \emph{Advances in Neural Information
  Processing Systems}, volume~38, 120957--120985. Curran Associates, Inc.

\bibitem[{Xi et~al.(2026)Xi, Lin, Xiao, Zhou, Shan, Gao, Zhu, Liu, Yu, and
  Zhang}]{xi2026searchagents}
Xi, Y.; Lin, J.; Xiao, Y.; Zhou, Z.; Shan, R.; Gao, T.; Zhu, J.; Liu, W.; Yu,
  Y.; and Zhang, W. 2026.
\newblock A Survey of Large Language Model-Based Search Agents.
\newblock In \emph{Proceedings of the 64th Annual Meeting of the Association
  for Computational Linguistics (Volume 1: Long Papers)}, 8244--8279.

\bibitem[{Xi et~al.(2025)Xi, Lin, Zhu, Xiao, Ou, Liu, Wan, Chen, Liu, Wang,
  Tang, Zhang, and Yu}]{xi2025infodeepseek}
Xi, Y.; Lin, J.; Zhu, M.; Xiao, Y.; Ou, Z.; Liu, J.; Wan, T.; Chen, B.; Liu,
  W.; Wang, Y.; Tang, R.; Zhang, W.; and Yu, Y. 2025.
\newblock {InfoDeepSeek}: Benchmarking Agentic Information Seeking for
  Retrieval-Augmented Generation.
\newblock \emph{arXiv preprint arXiv:2505.15872}.

\bibitem[{Xiao et~al.(2023)Xiao, Liu, Zhang, Muennighoff, Lian, and
  Nie}]{xiao2023cpack}
Xiao, S.; Liu, Z.; Zhang, P.; Muennighoff, N.; Lian, D.; and Nie, J.-Y. 2023.
\newblock {C-Pack}: Packed Resources for General Chinese Embeddings.
\newblock ArXiv:2309.07597.

\bibitem[{Xiong et~al.(2021)Xiong, Li, Iyer, Du, Lewis, Wang, Mehdad, Yih,
  Riedel, Kiela, and O{\u{g}}uz}]{xiong2021mdr}
Xiong, W.; Li, X.~L.; Iyer, S.; Du, J.; Lewis, P.; Wang, W.~Y.; Mehdad, Y.;
  Yih, W.-t.; Riedel, S.; Kiela, D.; and O{\u{g}}uz, B. 2021.
\newblock Answering Complex Open-Domain Questions with Multi-Hop Dense
  Retrieval.
\newblock In \emph{International Conference on Learning Representations}.

\bibitem[{Xu, Shi, and Choi(2024)}]{xu2024recomp}
Xu, F.; Shi, W.; and Choi, E. 2024.
\newblock {RECOMP}: Improving Retrieval-Augmented {LM}s with Context
  Compression and Selective Augmentation.
\newblock In \emph{International Conference on Learning Representations},
  volume 2024, 43478--43502.

\bibitem[{Yan et~al.(2024)Yan, Gu, Zhu, and Ling}]{yan2024crag}
Yan, S.-Q.; Gu, J.-C.; Zhu, Y.; and Ling, Z.-H. 2024.
\newblock Corrective Retrieval Augmented Generation.
\newblock ArXiv:2401.15884.

\bibitem[{Yang et~al.(2025)Yang, Zeng, Rao, and Zhang}]{yang2025simrag}
Yang, D.; Zeng, L.; Rao, J.; and Zhang, Y. 2025.
\newblock Knowing You Don{'}t Know: Learning When to Continue Search in
  Multi-round {RAG} through Self-Practicing.
\newblock In \emph{Proceedings of the 48th International ACM SIGIR Conference
  on Research and Development in Information Retrieval}, 1305--1315. ACM.

\bibitem[{Yang et~al.(2018)Yang, Qi, Zhang, Bengio, Cohen, Salakhutdinov, and
  Manning}]{yang2018hotpotqa}
Yang, Z.; Qi, P.; Zhang, S.; Bengio, Y.; Cohen, W.~W.; Salakhutdinov, R.; and
  Manning, C.~D. 2018.
\newblock {HotpotQA}: A Dataset for Diverse, Explainable Multi-Hop Question
  Answering.
\newblock In \emph{Proceedings of the 2018 Conference on Empirical Methods in
  Natural Language Processing}, 2369--2380.

\bibitem[{Yao et~al.(2023)Yao, Zhao, Yu, Du, Shafran, Narasimhan, and
  Cao}]{yao2023react}
Yao, S.; Zhao, J.; Yu, D.; Du, N.; Shafran, I.; Narasimhan, K.; and Cao, Y.
  2023.
\newblock {ReAct}: Synergizing Reasoning and Acting in Language Models.
\newblock In \emph{International Conference on Learning Representations}.

\bibitem[{Yao et~al.(2025)Yao, Qi, Pan, Cao, Hu, Liu, Hou, and
  Li}]{yao2025seakr}
Yao, Z.; Qi, W.; Pan, L.; Cao, S.; Hu, L.; Liu, W.; Hou, L.; and Li, J. 2025.
\newblock {SeaKR}: Self-aware Knowledge Retrieval for Adaptive Retrieval
  Augmented Generation.
\newblock In \emph{Proceedings of the 63rd Annual Meeting of the Association
  for Computational Linguistics (Volume 1: Long Papers)}, 27022--27043. Vienna,
  Austria: Association for Computational Linguistics.

\bibitem[{Yu et~al.(2026)Yu, Zhou, Xu, Guo, Shan, Fu, Wang, Liu, Yu, Zhang, and
  Lin}]{yu2026latentskill}
Yu, A.; Zhou, C.; Xu, T.; Guo, Z.; Shan, R.; Fu, Z.; Wang, J.; Liu, W.; Yu, Y.;
  Zhang, W.; and Lin, J. 2026.
\newblock {LatentSkill}: From In-Context Textual Skills to In-Weight Latent
  Skills for {LLM} Agents.
\newblock \emph{arXiv preprint arXiv:2606.06087}.

\bibitem[{Zhang et~al.(2024)Zhang, Liao, Li, Du, and Lin}]{zhang2024agenticir}
Zhang, W.; Liao, J.; Li, N.; Du, K.; and Lin, J. 2024.
\newblock Agentic Information Retrieval.
\newblock \emph{arXiv preprint arXiv:2410.09713}.

\bibitem[{Zhao et~al.(2026)Zhao, Wang, Xu, Zha, Bowen, Wang, Jia, Liu, and
  Wang}]{zhao2026rsearch}
Zhao, Q.; Wang, R.; Xu, D.; Zha, D.; Bowen, M.; Wang, Z.; Jia, S.; Liu, L.; and
  Wang, X. 2026.
\newblock {R-Search}: Empowering {LLM} Reasoning with Search via Multi-Reward
  Reinforcement Learning.
\newblock In \emph{Findings of the Association for Computational Linguistics:
  ACL 2026}, 38030--38046. San Diego, California, United States: Association
  for Computational Linguistics.

\bibitem[{Zheng et~al.(2025)Zheng, An, Wang, Wang, and
  Wu}]{zheng2025stepsearch}
Zheng, X.; An, K.; Wang, Z.; Wang, Y.; and Wu, Y. 2025.
\newblock {StepSearch}: Igniting {LLM}s Search Ability via Step-Wise Proximal
  Policy Optimization.
\newblock In \emph{Proceedings of the 2025 Conference on Empirical Methods in
  Natural Language Processing}, 21805--21830. Suzhou, China: Association for
  Computational Linguistics.

\bibitem[{Zhou et~al.(2026)Zhou, Chai, Chen, Guo, Shan, Song, Xu, Yang, Yu,
  Zhang, Zheng, Zhu, Zheng, Zhang, Lou, Zhang, Fu, Wang, Liu, Lin, and
  Zhang}]{zhou2026externalization}
Zhou, C.; Chai, H.; Chen, W.; Guo, Z.; Shan, R.; Song, Y.; Xu, T.; Yang, Y.;
  Yu, A.; Zhang, W.; Zheng, C.; Zhu, J.; Zheng, Z.; Zhang, Z.; Lou, X.; Zhang,
  C.; Fu, Z.; Wang, J.; Liu, W.; Lin, J.; and Zhang, W. 2026.
\newblock Externalization in {LLM} Agents: A Unified Review of Memory, Skills,
  Protocols and Harness Engineering.
\newblock \emph{arXiv preprint arXiv:2604.08224}.

\bibitem[{Zhuang et~al.(2024)Zhuang, Zhang, Cheng, Yang, Liu, Huang, Lin,
  Rajmohan, Zhang, and Zhang}]{zhuang2024efficientrag}
Zhuang, Z.; Zhang, Z.; Cheng, S.; Yang, F.; Liu, J.; Huang, S.; Lin, Q.;
  Rajmohan, S.; Zhang, D.; and Zhang, Q. 2024.
\newblock {EfficientRAG}: Efficient Retriever for Multi-Hop Question Answering.
\newblock In \emph{Proceedings of the 2024 Conference on Empirical Methods in
  Natural Language Processing}, 3392--3411. Miami, Florida, USA: Association
  for Computational Linguistics.

\end{thebibliography}
